\documentclass[conference]{IEEEtran}
\IEEEoverridecommandlockouts
\usepackage{cite}
\usepackage{amsmath,amssymb,amsfonts}
\usepackage{algorithmic}
\usepackage{graphicx}
\usepackage{textcomp}
\usepackage{xcolor}

\usepackage{booktabs}
\usepackage{soul}
\usepackage{url}

\usepackage{tikz}
\usepackage{textcomp}
\usepackage{hyperref}
\usepackage{lipsum}
\newcommand\copyrighttext{%
  \footnotesize \textcopyright 2019 IEEE. Personal use of this material is permitted.
  Permission from IEEE must be obtained for all other uses, in any current or future
  media, including reprinting/republishing this material for advertising or promotional
  purposes, creating new collective works, for resale or redistribution to servers or
  lists, or reuse of any copyrighted component of this work in other works.
  DOI: \href{<http://tex.stackexchange.com>}{10.1109/BigData47090.2019.9005672}}
\newcommand\copyrightnotice{%
\begin{tikzpicture}[remember picture,overlay]
\node[anchor=south,yshift=10pt] at (current page.south) {\fbox{\parbox{\dimexpr\textwidth-\fboxsep-\fboxrule\relax}{\copyrighttext}}};
\end{tikzpicture}%
}

\makeatletter
\let\MYcaption\@makecaption
\makeatother

\usepackage[font=footnotesize]{subcaption}

\makeatletter
\let\@makecaption\MYcaption
\makeatother

\def\BibTeX{{\rm B\kern-.05em{\sc i\kern-.025em b}\kern-.08em
    T\kern-.1667em\lower.7ex\hbox{E}\kern-.125emX}}
\begin{document}

\title{Mixture-based Multiple Imputation Model for Clinical Data with a Temporal Dimension\\
\thanks{This work is supported in part by NIH grant R21LM012618.}
}

\author{\IEEEauthorblockN{1\textsuperscript{st} Ye Xue}
\IEEEauthorblockA{
\textit{Northwester University}\\
Evanston, USA \\
ye.xue@u.northwestern.edu}
\and
\IEEEauthorblockN{2\textsuperscript{nd} Diego Klabjan}
\IEEEauthorblockA{
\textit{Northwestern University}\\
Evanston, USA \\
d-klabjan@northwestern.edu}
\and
\IEEEauthorblockN{3\textsuperscript{rd} Yuan Luo}
\IEEEauthorblockA{
\textit{Northwestern University}\\
Chicago, USA \\
yuan.luo@northwestern.edu}
}

\maketitle
\copyrightnotice

\begin{abstract}
The problem of missing values in multivariable time series is a key challenge in many applications such as clinical data mining. Although many imputation methods show their effectiveness in many applications, few of them are designed to accommodate clinical multivariable time series. In this work, we propose a multiple imputation model that capture both cross-sectional information and temporal correlations. We integrate Gaussian processes with mixture models and introduce individualized mixing weights to handle the variance of predictive confidence of Gaussian process models. The proposed model is compared with several state-of-the-art imputation algorithms on both real-world and synthetic datasets. Experiments show that our best model can provide more accurate imputation than the benchmarks on all of our datasets.
\end{abstract}


\section{Introduction}

The computational modeling in clinical applications attracts growing interest with the realization that the quantitative understanding of patient pathophysiological progression is crucial to clinical studies \cite{winslow2012computational}. With a comprehensive and precise modeling, we can have a better understanding of a patient's state, offer more precise diagnosis and provide better individualized therapies \cite{kohane2015ten}. Researchers are increasingly motivated to build more accurate computational models from multiple types of clinical data. However, missing values in clinical data challenge researchers using analytic techniques for modeling, as many of the techniques are designed for complete data.

Traditional strategies used in clinical studies to handle missing values include deleting records with missing values and imputing missing entries by mean values. However, deleting records with missing values and some other filtering strategies can introduce biases \cite{weber2017biases} that can impact modeling in many ways, thus limiting its generalizability. Mean imputation is widely used by researchers to handle missing values. However, it is shown to yield less effective estimates than many other modern imputation techniques \cite{brown1994efficacy,wothke2000longitudinal,arbuckle1996full,waljee2013comparison}, such as maximum likelihood approaches and multiple imputation methods (e.g. multiple imputation by chained equations (MICE) \cite{buuren2011mice}). The maximum likelihood and multiple imputation methods are based on solid statistical foundations and become standard in the last few decades \cite{graham2009missing,schafer2002missing}.

Multiple imputation is originally proposed by Rubin \cite{rubin1978multiple} and with the idea of replacing each missing value with multiple estimates that reflect variation within each imputation model. The classic multiple imputation method treats the complete variables as predictors and incomplete variables as outcomes, and uses multivariate regression models to estimate missing values \cite{rubin1987multiple,schafer1997analysis}. Another imputation approach relies on chained equations where a sequence of regression models is fitted by treating one incomplete variable as the outcome and other variables with possible best estimates for missing values as predictors \cite{raghunathan2001multivariate,van2018flexible,van2006fully}. Multiple imputation has its success in healthcare applications, where many imputation methods designed for various tasks are based on multiple imputation \cite{harel2007multiple,qi2010comparison,su2011multiple,hsu2006survival,long2012doubly,van1999multiple,deng2016multiple}.

In recent years, additional imputation methods are proposed. Although many imputation methods \cite{waljee2013comparison,buuren2011mice,stekhoven2011missforest,raghunathan2001multivariate,little2004robust,luo2016using,zhang2009extensions,troyanskaya2001missing} show their effectiveness in many applications, few of them are designed for time series-based clinical data. These clinical data are usually multivariable time series, where patients have measurements of multiple laboratory tests at different times. Many methods are designed for cross-sectional imputation (measurements taken at the same time point) and do not consider temporal information that is useful in making predictions or imputing missing values. Ignoring informative temporal correlations and only capturing cross-sectional information may yield less effective imputation. 

In order to address the limitations mentioned above, we present a mixture-based multiple imputation model (MixMI) for clinical time series \footnote{ The implementations of our model is available at: \url{https://github.com/y-xue/MixMI}.}. Our model captures both cross-sectional information and temporal correlations to estimate missing values using mixture models. We model the distribution of measurements using a mixture model. The mixture is composed of linear regression to model cross-sectional correlations and Gaussian processes (GPs) to capture temporal correlations. The problem of integrating GP within a standard mixture model is that GP models in all patient cases get the same mixing weights, while the confidence of predictions by GP models can vary largely across different patient cases. We overcome this problem by introducing individualized mixing weights for each patient cases, instead of assigning a fixed weight. We train our model using the Expectation-Maximization (EM) algorithm. We demonstrate the effectiveness of our model by comparing it with several state-of-the-art imputation algorithms on multiple clinical datasets.

Our main contributions are summarized as follows.

1. To the best of our knowledge, we are the first to build imputation model for time series by integrating GP within mixture models. We address the problem that all GP models in all patient cases get a fixed mixing weight by introducing individualized mixing weights.

2. We test the performance of our model on two real-world clinical datasets and several synthetic datasets. Our best model outperforms all comparison models including several state-of-the-art imputation models. Using synthetic datasets, we also explore and discover the properties of the data that benefit our model and/or comparison models. Experiments show that our best model is robust to the variation of these properties and outperforms comparison models on all synthetic datasets.

The remainder of this paper is structured as follows. Section \ref{RW} discusses related work while in Section \ref{METHOD}, the proposed method is described. The experimental setup, including dataset collection and evaluation procedure, is described in Section \ref{ES}. Section \ref{RES} discusses the computational results and underlying analyses. The conclusions are drawn in Section \ref{CONCLUDE}.

\section{Related work} \label{RW}

Research in designing imputation methods for multivariable time series attracts growing interest in recent decades. Previous studies generally fall into two categories. One comes from methods using linear or other simple parametric functions to estimate missing values. The other is the methods treating time series as smooth curves and estimating missing values using GP or other nonparametric methods. 

In the first category, multivariable time series are modeled based on either linear models \cite{shah1997random}, linear mixed models \cite{liu2000multiple,schafer2002computational} or autoregressive models \cite{holmes2012marss,bashir2017handling}. However, in these methods, the potential trajectories of variables are only limited to linear or other simple parametric functions. Alternatively, many authors choose GPs or other nonparametric functions to model time series. Compared to linear models, GPs only have locality constraints in which close time points in a time series usually have close values. Therefore, GPs bring in more flexibility in capturing temporal trajectories of variables.

However, directly applying GPs on imputation for multivariable time series usually yields less effective imputation due to the lack of considerations for the similarities and correlations across time series. Futoma et al. \cite{futoma2017learning} extends the GP-based approach to imputation in the multivariable setting by applying the multi-task Gaussian Processes (MTGP) prediction \cite{bonilla2008multi}. However, the quality of estimating missing values relies on the estimation of covariance structure among variables when using MTGP or other multi-task functional approaches \cite{he2011functional,chiou2014functional,kliethermes2014bayesian}. To make a confident estimation of the covariance, a large amount of time points with shared observations of these variables are often required by these multi-task approaches \cite{bonilla2008multi,yu2005learning}. Due to the fact that many patients only have records with a limited number of time points, time series of inpatient clinical laboratory tests fall short of such a requirement. Hori et al. \cite{hori2016multi} extend MTGP to imputation on multi-environmental trial data in a third-order tensor. However, this approach is not applicable to clinical data with a large number of patients, since the covariance matrix of observed values needs to be computed together with its inverse, which is intractable for our datasets.

Recently, Luo et al. \cite{luo20173d} explore the application of GPs in clinical data and propose an algorithm, 3-dimensional multiple imputation with chained equations (3D-MICE), that combines the imputation from GP and traditional MICE based on weighting equations. However, the weighting equations are calculated only based on the standard deviations of GP and MICE predictions for missing values. The weighting strategy is static and not optimized. We postulate that calculating weights through an optimization problem can help to improve the imputation quality. In our work, instead of the predictive mean matching used in \cite{luo20173d}, we choose linear regression as one component of our model. Our model is also based on a statistical model and thus statistically justified which is not the case for \cite{luo20173d}. 

The mixture model proposed in this work differs from the mixtures of Gaussian processes (MGP) \cite{tresp2001mixtures} in several ways. The mixing components in our model are not limited to Gaussian process, thus bringing in possibilities of capturing correlations with various choices of parametric functions. More importantly, our model is designed for multivariable time series and captures both temporal and cross-sectional correlations. The latter is ignored in \cite{tresp2001mixtures} when MGP is applied to multivariable time series, which yields less effective predictions since some clinical tests have strong cross-sectional relations. Without a Gaussian process component, the standard Gaussian mixture model (GMM) is well studied and has been extended to imputation tasks \cite{di2007imputation,delalleau2012efficient,yan2015missing,silva2018multivariate}. These approaches still suffer from losing informative temporal correlations when imputing missing values in longitudinal data.

In order to effectively model the interaction between different aspects, we represent the data as a tensor with each aspect being one mode. Therefore, our method is also considered as a tensor completion approach. Tensor completion problems are extensively studied. However, the classic tensor completion methods \cite{filipovic2015tucker,liu2015trace,tomioka2010estimation} focus on general tensors and usually do not consider temporal aspects. In recent years, many studies explore the application of temporal augmented tensor completion on imputing missing values in time series \cite{bahadori2014fast,yu2015accelerated,ge2016uncovering,takeuchi2017autoregressive,cai2015facets}. These methods discretize time into evenly sampled intervals. However, due to the fact that inpatient clinical laboratory tests are usually measured at varying intervals, assembling clinical data over regularly sampled time periods might have several drawbacks, such as leading to sparse tensors if discretizing time at fine granularity (e.g. every minute) while some laboratory tests are measured less frequently (e.g. daily). Furthermore, extending these methods to the case, where time is not regularly sampled, is not easy and straightforward, requiring changing design details and the objective functions to be optimized. 
Recently, Yang et al. \cite{yang2018time} propose a tensor completion method that can deal with irregularly sampled time. However, most components of this approach are tailored to the characteristics of septic patients. We implement this approach with only the component of time-aware mechanism that is general and applicable to our experimental settings. However, this approach is not so effective in our experiments, thus not included as a benchmark in this paper. Lately, Zhe et al. \cite{zhe2018stochastic} propose a Bayesian nonparametric tensor decomposition model that captures temporal correlations between interacting events. However, this approach is not directly applicable to continuous multivariable time series because it focuses on discrete events and captures temporal correlations between the occurrences of events. 

Recurrent Neural Networks (RNNs) are another choice to capture temporal relationships and have been studied in imputation tasks. Early studies \cite{bengio1996recurrent,tresp1998solution,parveen2002speech} in RNN-based imputation models capture temporal correlations either within each time series or across all time series \cite{yoon2018estimating}. Yoon et al. \cite{yoon2018estimating} recently propose a novel multi-directional RNN that captures temporal correlations both within and across time series. Compared with traditional statistical models, neural network-based models are usually harder to train and provide limited explainability, while the performance is not always necessarily better. We use \cite{yoon2018estimating} as a benchmark since it is the most recent model. Besides imputation, researchers also attempt to build classification models with RNNs that utilize the missing data patterns \cite{choi2016doctor,lipton2016directly,che2018recurrent}. Instead of relying on the estimation of missing values, these models perform predictions by exploring the missingness itself and no separate imputation is required. These models can not be a substitute for imputation methods since they do not provide imputed values.


\section{Methodology} \label{METHOD}

\subsection{Imputation Framework}

\begin{figure}[htbp]
  \centering
  \includegraphics[width=0.9\linewidth]{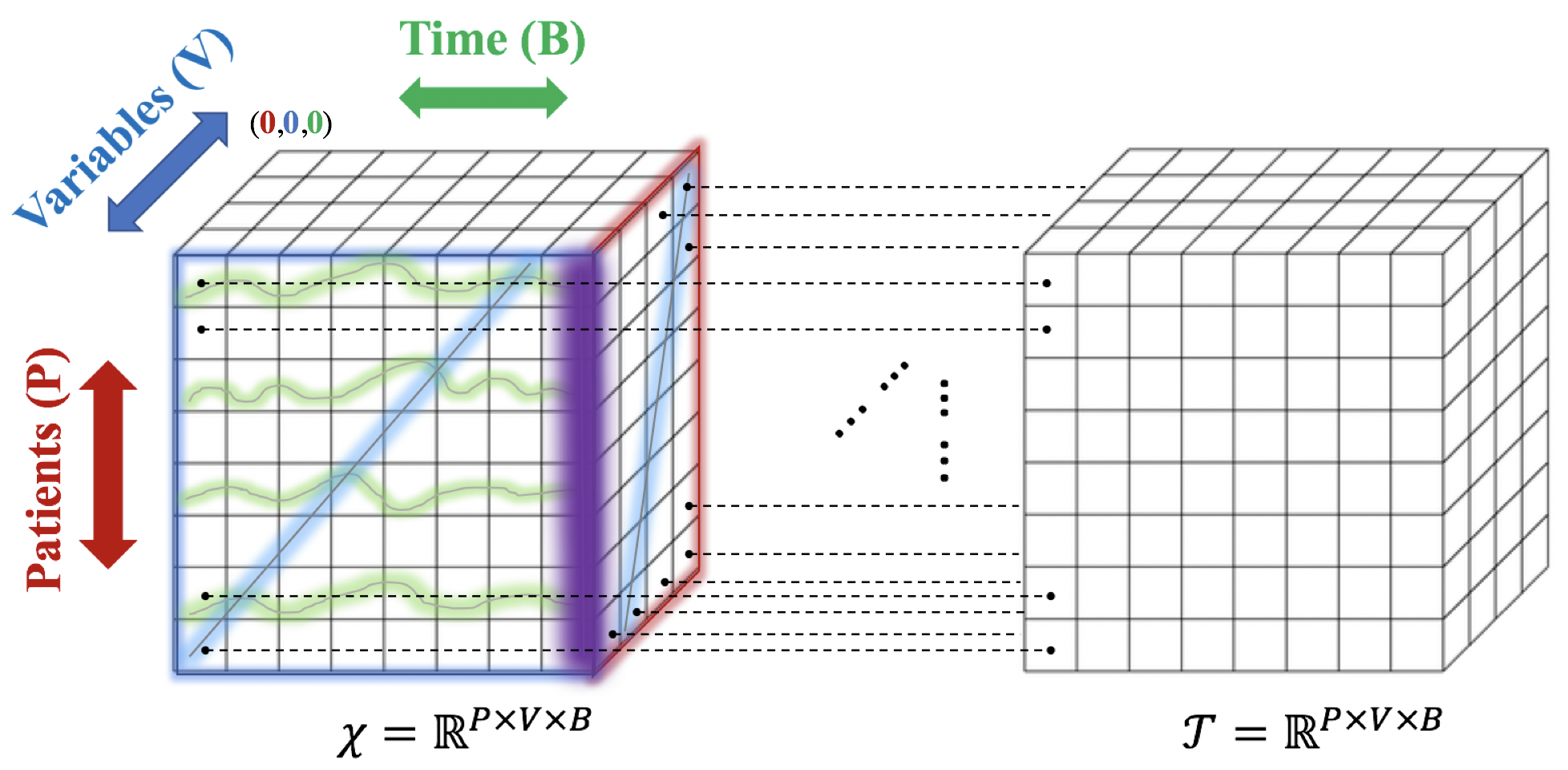}
  \caption{Measurement and time tensor. An example of the inputs and output of the mixture model $Mix_{V,B}$ is shown as the colorful fiber and matrices. In $Mix_{V,B}$, the target output is $x_{:,V,B}$ shown in purple and the inputs are $x_{:,-V,B}$ and $x_{:,V,-B}$ shown in red and blue matrices, respectively, excluding the purple fiber. We model the output with a mixture model, where we train a linear regression on the red matrix and train GPs or/and another linear regression on the blue matrix.}
  \label{fig:tensor}
\end{figure}

In many predictive tasks on temporal clinical data, time series are often aligned into the same-length sequences to derive more robust patient phenotypes through matrix decomposition or discover feature groups by applying sequence/graph mining techniques \cite{yang2018time,xue2018predicting}. We model under this assumption. In this work, we use tensor representation, in which patients have the same number of time points. We represent the data collected from $P$ patients with $V$ laboratory tests and $B$ time points as two 3D tensors $\mathcal{X} \in \mathbb{R}^{P \times V \times B}$ and $\mathcal{T} \in \mathbb{R}^{P \times V \times B}$, shown in Fig.~\ref{fig:tensor}. Each laboratory test measurement $x_{p,v,b}$ is stored in the measurement tensor $\mathcal{X}$. Each time $t_{p,v,b}$, when $x_{p,v,b}$ is measured, is stored in the time tensor $\mathcal{T}$. 

Table~\ref{tab:symbol} lists the main symbols we use throughout the paper. Missing values in the measurement tensor are denoted as $x^{mis}_{p,v,b}$, showing that the value of test $v$ at time index $b$ for patient $p$ is missing. Correspondingly, $x^{obs}_{p,v,b}$ denotes an observed value. The time tensor $\mathcal{T}$ is complete, since we only collect patient records at the time when at least one laboratory test measurement is available. We assume we know the ``prescribed'' time a missing measurement should have been taken. In the matrix $x_{:,:,b}$ at time index $b$, the measurement time $t_{p,v,b}$ and $t_{q,v,b}$ can be different when $p \neq q$, whereas for a given patient $p$, we have $t_{p,v,b} = t_{p,u,b}$ for $v,u \in [1:V]$. That is, all tests for a particular patient are taken at the same time. 

Disregarding the temporal dimension, the imputation problem is well studied. If one dimension is time, in order to apply imputation methods that are not designed for time series, we need to disregard the temporal aspect or ignore temporal correlations of the data. However, temporal trajectories can reveal patient's underlying pathophysiological evolution, the modeling of which can help to better estimate missing values. For the reason that both cross-sectional information and temporal aspects can impact the estimation of missing measurements, we explore mixture models, which are composed of several base models through either a cross-sectional or temporal view. We introduce these base models in Section \ref{BM}.

\begin{table}
  \caption{Main symbols and definitions}
  \begin{tabular}[t]{ll}
    \toprule
    Symbol&Definition \\
    \midrule
    $\mathcal{X}$, $\mathcal{T}$ & Measurement and time tensor  \\
    $x_{p,v,-b}$ & Measurements in fiber $x_{p,v,:}$ excluding $x_{p,v,b}$ \\
    $t_{p,v,-b}$ & Times in fiber $t_{p,v,:}$ excluding $t_{p,v,b}$ \\
    $Mix_{v,b}$ & Mixture model for $v$ and $b$. \\
    $V_{v,b}$ & Concatenation of $x_{:,-v,b}$ and $x_{:,v,-b}$ \\
    $\mathcal{N}$($\boldsymbol{\mu}^{(k)}_{v,b}$,$\boldsymbol{\Sigma}^{(k)}_{v,b}$) & \parbox[t]{6cm}{The $k$th prior multivariate normal distribution in $Mix_{v,b}$} \\
    $m^{(3)}(\cdot)$, $\Sigma^{(3)}(\cdot)$ & Predictive mean and variance of a GP model \\
    $\gamma_{v,b}$ & \parbox[t]{6cm}{The set of all trainable parameters of $Mix_{v,b}$}\\
    $I$ & The identity matrix\\
  \bottomrule
\end{tabular}
\label{tab:symbol}
\end{table}

In our imputation framework, a mixture model is trained for each variable and time index. We use $Mix_{v,b}$ to denote the mixture model to impute missing values of variable $v$ at time index $b$. The missing values $x^{mis}_{:,v,b}$ in the fiber $x_{:,v,b}$ are imputed by the optimized $Mix_{v,b}$. In each iteration of the algorithm, it is assumed that all other values are known and only $x^{mis}_{:,v,b}$ is imputed. This is wrapped in an outer loop. We call this procedure simple-pass tensor imputation, i.e., one pass through all $v,b$. Since several simple-pass tensor imputations are conducted, our approach is also considered as an iterative imputation \cite{gelman2004parameterization}, which can also be regarded as a sampling-based approach where a Gibbs sampler is used to approximate convergence. The convergence of iterative imputation methods can be quite fast with a few iterations \cite{buuren2011mice}. 

In detail, the iterative imputation approaches start by replacing all missing data with values from simple guesses; we fill in all missing values with initial estimates by taking random draws from observed values. This procedure is called an initial imputation. Then we perform iterative tensor imputation on each copy separately. The training procedure and imputation for fibers are introduced in Section \ref{MM} and \ref{MPE}. We also rely on the concept of multiple imputations, where several iterative imputations are performed and the imputed values are averaged at the end. Each iterative imputation starts with a different iterative imputation tensor and/or uses a different order of $v$,$b$.

In summary, the algorithm creates $M$ different copies $\mathcal{X}^1, \ldots, \mathcal{X}^M$ of $\mathcal{X}$, each one filled with different random $\mathcal{X}^{mis}$. For each $i=1, \ldots, M$, we then perform $K$ simple-pass tensor imputations. Each simple-pass has a loop over all $v$,$b$, which uses $Mix_{v,b}$ to adjust $\mathcal{X}^{i,mis}$. At the end of the $M$ imputations, $\mathcal{X}^{i,mis}$ are averaged across all $i$ to yield the final imputed tensor. 

The whole imputation process involves $M \times K \times V \times B$ imputation models. We next first focus on the base models behind $Mix_{v,b}$ and then on the actual mixture model. 

\subsection{Base Models} \label{BM}

Our mixture models are composed of three components that are derived from two base models, linear regression and Gaussian processes. One component consists of GP models and the other two components are linear models through two different views of the measurement tensor. Through a cross-sectional view, the tensor can be considered as a vector of patient-by-variable matrices at different time indices. Through a temporal view, we can view the tensor as a vector of patient-by-time matrices for different variables. 

\subsubsection{Linear model through cross-sectional view}

We can view the measurement tensor $\mathcal{X}$ as a vector of patient-by-variable matrices. On the slice $x_{:,:,b}$, we use a linear regression model to fit the target variable $v$ as a function of the other variables except $v$. The target values $x_{:,v,b}$ are modeled as

\begin{equation}
x_{:,v,b} = x_{:,-v,b} \beta^{(1)}_{v,b} + \epsilon^{(1)}_{v,b}, \epsilon^{(1)}_{v,b} \sim \mathcal{N} (0,{\sigma^{(1)}_{v,b}}^2 I) \label{eq1}
\end{equation}
where $\beta^{(1)}_{v,b}$ is the column vector of coefficients and $\sigma^{(1)}_{v,b}$ is the standard deviation of the error $\epsilon^{(1)}_{v,b}$, regarding the regression model through cross-sectional view for variable $v$ and time index $b$.

The likelihood distribution of $x_{:,v,b}$ is then given by
\begin{equation}
x_{:,v,b}|x_{:,-v,b},\beta^{(1)}_{v,b},{\sigma^{(1)}_{v,b}}^2 \sim \mathcal{N} (x_{:,-v,b}\beta^{(1)}_{v,b},{\sigma^{(1)}_{v,b}}^2 I).
\end{equation}

The training data consists of observed target values $(x_{p,v,b})_{p \in P^{tr}_{v,b}}$ and the input data $(x_{p,-v,b})_{p \in P^{tr}_{v,b}}$, where $P^{tr}_{v,b}$ is the training patient set and includes $p$ only if $x_{p,v,b}$ is observed.

\subsubsection{Linear model through temporal view}

In addition to the cross-sectional view, we can also view the measurement tensor $\mathcal{X}$ as a vector of patient-by-time matrices. On matrix $x_{:,v,:}$, we use linear regression to model the measurements at time index $b$ against those at other indices.

The target values $x_{:,v,b}$ are modeled as
\begin{equation}
x_{:,v,b} = x_{:,v,-b} \beta^{(2)}_{v,b} + \epsilon^{(2)}_{v,b}, \epsilon^{(2)}_{v,b} \sim \mathcal{N} (0,{\sigma^{(2)}_{v,b}}^2 I)
\end{equation}
where $\beta^{(2)}_{v,b}$ is the column vector of coefficients and $\sigma^{(2)}_{v,b}$ is the standard deviation of the error $\epsilon^{(2)}_{v,b}$, regarding the linear regression model though temporal view for variable $v$ and time index $b$.

The likelihood distribution of $x_{:,v,b}$ is given by
\begin{equation}
x_{:,v,b}|x_{:,v,-b},\beta^{(2)}_{v,b},{\sigma^{(2)}_{v,b}}^2 \sim \mathcal{N} (x_{:,v,-b}\beta^{(2)}_{v,b},{\sigma^{(2)}_{v,b}}^2 I).
\end{equation}

The training data consists of observed target values $(x_{p,v,b})_{p \in P^{tr}_{v,b}}$ and input data $(x_{p,v,-b})_{p \in P^{tr}_{v,b}}$. 

\subsubsection{Gaussian processes through temporal view}

Gaussian processes are commonly used to capture trajectories of variables, thus used in our mixture model to capture temporal correlations. Through the same temporal view as introduced above, on matrix $x_{:,v,:}$, we fit GPs on time series for each patient.

The target value $x_{p,v,b}$ is modeled as 
\begin{equation}
\begin{split}
x_{p,v,b} &= \mu_{p,v,b} + f(t_{p,v,b}), \\
f(t_{p,v,b}) &\sim \mathcal{GP} (0,\mathcal{K}(t_{p,v,b},t_{p,v,b'}))
\end{split}
\end{equation}
where $\mu_{p,v,b}$ is the overall mean of the model, $f(\cdot)$ is a Gaussian process with mean of $0$ and a covariance matrix $\mathcal{K}(t,t')$ of time pairs ($t$,$t'$). Then the likelihood distribution of $x_{p,v,b}$ is written as
\begin{equation}
\begin{split}
x_{p,v,b} | \alpha_{p,v,b} &\sim \mathcal{N} (m^{(3)}(\alpha_{p,v,b}), \Sigma^{(3)}(\alpha_{p,v,b})) \\
\alpha_{p,v,b} &= (\theta_{v,b}, x_{p,v,-b}, t_{p,v,-b}) 
\end{split}
\end{equation}
where $\theta_{v,b}$ are the kernel parameters of the GP models, and the predictive mean and variance are given by $m^{(3)}(\cdot)$ and $\Sigma^{(3)}(\cdot)$; see more details in Appendix B. For a certain $v$ and $b$, all GP models share the same kernel parameters $\theta_{v,b}$. 

\subsection{The Mixture Model} \label{MM}

Given the likelihood distribution of all three components, we model the joint mixture distribution, regarding the variable $v$ and time index $b$, in the following way
\begin{equation}
\begin{split}
p(&x_{:,v,b},V_{v,b}) \\ = &\pi^{(1)}_{v,b} \mathcal{N} (V_{v,b}|\delta^{(1)}_{v,b}) \mathcal{N}(x_{:,v,b}|x_{:,-v,b}\beta^{(1)}_{v,b},{\sigma^{(1)}_{v,b}}^2I) \\ + 
&\pi^{(2)}_{v,b} \mathcal{N} (V_{v,b}|\delta^{(2)}_{v,b}) \mathcal{N}(x_{:,v,b}|x_{:,v,-b}\beta^{(2)}_{v,b},{\sigma^{(2)}_{v,b}}^2I) \\ +
&\pi^{(3)}_{v,b} \mathcal{N} (V_{v,b}|\delta^{(3)}_{v,b}) \mathcal{N}(x_{:,v,b}|m^{(3)}(\boldsymbol{\alpha}_{v,b}),\text{diag}(\Sigma^{(3)}(\boldsymbol{\alpha}_{v,b})))
\end{split}
\end{equation}
where we define $V_{v,b} = [x_{:,-v,b}\ x_{:,v,-b}]$, $\boldsymbol{\alpha}_{v,b} = (\alpha_{p,v,b})_{p \in P}$, $\delta^{(k)}_{v,b} = (\boldsymbol{\mu}^{(k)}_{v,b},\boldsymbol{\Sigma}^{(k)}_{v,b})$ and $\pi^{(k)}_{v,b}$ is the mixing weight for the $k$th component. This model can be interpreted as the joint distribution between observed data $V_{v,b}$ and missing values $x_{:,v,b}$, which consists of a mixture of three distributions. The first one $p_1 (x_{:,v,b},V_{v,b})$ is modeled as 
\begin{equation}
\begin{split}
p_1 &(x_{:,v,b},V_{v,b}) \\
&= p(x_{:,v,b}|V_{v,b})p(V_{v,b}) \\
&= \mathcal{N} (x_{:,v,b}|\mu_{1}(V_{v,b}),\sigma_{1}(V_{v,b})) \mathcal{N} (V_{v,b}|\delta^{(1)}_{v,b}) \\
&= \mathcal{N} (x_{:,v,b}|x_{:,-v,b}\beta^{(1)}_{v,b},{\sigma^{(1)}_{v,b}}^2 I) \mathcal{N} (V_{v,b}|\delta^{(1)}_{v,b})
\end{split}
\end{equation} 
the remaining two follow the same logic.

By marginalizing over $x_{:,v,b}$, the prior probability distribution $p(V_{v,b})$ is written as
\begin{equation}
p(V_{v,b}) = \sum^{3}_{k=1} \pi^{(k)}_{v,b} \mathcal{N} (V_{v,b}|\delta^{(k)}_{v,b})
\end{equation}
which is a mixture of Gaussians. It also follows that
\begin{equation}
\begin{split}
p&(x_{:,v,b}|V_{v,b}) \\
&= \frac{p(x_{:,v,b},V_{v,b})}{p(V_{v,b})} \\
&= C^{(1)}_{v,b} \mathcal{N} (x_{:,v,b}|x_{:,-v,b}\beta^{(1)}_{v,b},{\sigma^{(1)}_{v,b}}^2I) \\
&+ C^{(2)}_{v,b} \mathcal{N} (x_{:,v,b}|x_{:,v,-b}\beta^{(2)}_{v,b},{\sigma^{(2)}_{v,b}}^2I) \\
&+ C^{(3)}_{v,b} \mathcal{N} (x_{:,v,b}|m^{(3)}(\boldsymbol{\alpha}_{v,b}),\text{diag}(\Sigma^{(3)}(\boldsymbol{\alpha}_{v,b})))
\end{split}
\end{equation}
where we define $C^{(k)}_{v,b} = \frac{\pi^{(k)}_{v,b} \mathcal{N} (V_{v,b}|\delta^{(k)}_{v,b})}{\sum_{j=1}^{3} \pi^{(j)}_{v,b} \mathcal{N} (V_{v,b}|\delta^{(j)}_{v,b})}$.

We train our mixture model on observed target values by maximizing the log likelihood of the joint mixture distribution
\begin{displaymath}
\hat{\gamma}_{v,b} = \arg \max_{\gamma_{v,b}} \ln{p(x^{obs}_{:,v,b},V^{tr}_{v,b};\gamma_{v,b})}
\end{displaymath}
where $V^{tr}_{v,b}$ is the training input data and defined as the concatenation of $(x_{p,-v,b})_{p \in P^{tr}_{v,b}}$ and $(x_{p,v,-b})_{p \in P^{tr}_{v,b}}$, and $\gamma_{v,b}$ is the set of all trainable parameters
\begin{displaymath}
\beta^{(1)}_{v,b},{\sigma^{(1)}_{v,b}}^2,\beta^{(2)}_{v,b},{\sigma^{(2)}_{v,b}}^2,\theta_{v,b}\ \text{and}\ \pi^{(k)}_{v,b},\boldsymbol{\mu}^{(k)}_{v,b},\boldsymbol{\Sigma}^{(k)}_{v,b}\ \text{for}\ k=1,2,3.
\end{displaymath}

After training, the missing values are imputed using individualized (per-patient) mixing weights that are derived from the conditional distribution. The missing value $x^{mis}_{p,v,b}$ is imputed by
\begin{displaymath}
\Pi^{(1)}_{p,v,b} x_{p,-v,b} \hat{\beta}^{(1)}_{v,b} + \Pi^{(2)}_{p,v,b} x_{p,v,-b} \hat{\beta}^{(2)}_{v,b} + \Pi^{(3)}_{p,v,b} m^{(3)}(\hat{\alpha}_{p,v,b})
\end{displaymath}
where the individualized mixing weight of the $k$th component for patient $p$, variable $v$ and time index $b$ is defined as
\begin{equation}
\Pi^{(k)}_{p,v,b} = \frac{\hat{\pi}^{(k)}_{v,b} \mathcal{N} (V_{p,v,b}|\boldsymbol{\hat{\mu}}^{(k)}_{v,b},\boldsymbol{\hat{\Sigma}}^{(k)}_{v,b})}{\sum_{j=1}^{3} \hat{\pi}^{(j)}_{v,b} \mathcal{N} (V_{p,v,b}|\boldsymbol{\hat{\mu}}^{(j)}_{v,b},\boldsymbol{\hat{\Sigma}}^{(j)}_{v,b})}, k = 1,2,3 \label{eq:PI}
\end{equation}
where $V_{p,v,b}$ is the observed data and defined as the concatenation of $x_{p,-v,b}$ and $x_{p,v,-b}$.

\subsection{Mixture Parameter Estimation} \label{MPE}

Let $\ell(\gamma_{v,b})$ be the log likelihood $\ln{p(x^{obs}_{:,v,b},V^{tr}_{v,b};\gamma_{v,b})}$. Explicitly maximizing $\ell(\gamma_{v,b})$ is hard. Instead, we use the EM algorithm to repeatedly construct a lower-bound on $\ell(\gamma_{v,b})$ and then optimize that lower-bound $\mathcal{L}(\gamma_{v,b})$. We first define a latent indicator variable $q_{v,b} \in \{1,2,3\}$ that specifies which mixing component that data points come from. Then we use Jensen's inequality to get the lower-bound $\mathcal{L}(\gamma_{v,b})$, which is given by
\begin{equation}
\begin{split}
\mathcal{L} &(\gamma_{v,b}) \\
&= \sum_{p \in P^{tr}_{v,b}} \sum_{q_{v,b}} Q_{p}(q_{v,b}) \ln \frac{p(x_{p,v,b},V_{p,v,b},q_{v,b};\gamma_{v,b})}{Q_{p}(q_{v,b})} \\
&\leq \sum_{p \in P^{tr}_{v,b}} \ln \sum_{k=1}^{3} p(q_{v,b}=k) p(x_{p,v,b},V_{p,v,b};\gamma_{v,b}|q_{v,b}=k) \\
&= \ell(\gamma_{v,b})
\end{split}
\label{eq:loss}
\end{equation}
where
\begin{equation}
Q_p(q_{v,b}=k) = \frac{p(q_{v,b}=k) p(x_{p,v,b},V_{p,v,b}|q_{v,b}=k)}{\sum_{j=1}^{3} p(q_{v,b}=j) p(x_{p,v,b},V_{p,v,b}|q_{v,b}=j)}. \label{eq:Q}
\end{equation}

In \eqref{eq:loss} and \eqref{eq:Q}, the marginal distribution $p(q_{v,b}=k)$ over $q_{v,b}$ is specified by the mixing coefficients $\pi^{(k)}_{v,b} = p(q_{v,b}=k)$. We can view $Q_p(q_{v,b}=k)$ as the responsibility that component $k$ of the mixture model $Mix_{v,b}$ takes to ``explain'' $x_{p,v,b}$. We use the standard EM algorithm to maximize the lower-bound $\mathcal{L}(\gamma_{v,b})$; see more details about the estimation of the parameters in Appendix A.

\begin{table*}
  \centering
  \caption{Overall MASE by dataset and imputation model. The bold numbers are the significantly best values among all imputation models.}
  \begin{tabular}{ccccccccc}
    \toprule
    Dataset&GP&MTGP$^{1}$&M-RNN&GMM&MICE&3D-MICE&MixMI-LL&MixMI\\
    \midrule
    Real-world MIMIC ($d$=0)    & 0.13072 & 0.12599 & 0.12512 & 0.11741 & 0.11763 & 0.11186 & 0.09304 & \bfseries 0.09285 \\ 
    Synthetic MIMIC ($d$=0.5)   & 0.10466 & -       & 0.12320 & 0.11455 & 0.11573 & 0.09087 & 0.08612 & \bfseries 0.08448 \\ 
    Synthetic MIMIC ($d$=1)     & 0.09220 & -       & 0.12201 & 0.11436 & 0.11561 & 0.07715 & 0.08427 & \bfseries 0.07538 \\ 
    NMEDW                       & 0.18353 & 0.17370 & 0.14607 & 0.13593 & 0.13718 & 0.13624 & 0.11600 & \bfseries 0.11589 \\
  \bottomrule
  \multicolumn{8}{l}{$^{1}$ MTGP cannot have multiple inputs, thus not applicable to the synthetic datasets.}
\end{tabular}
\label{tab:maseALL}
\end{table*}

\begin{table*}
  \centering
  \caption{MASE on the real-world MIMIC dataset by variable and imputation model. The bold numbers are the best values among all imputation models.}
  \begin{tabular}{ccccccccc}
    \toprule
    Variable&GP&MTGP&M-RNN&GMM&MICE&3D-MICE&MixMI-LL&MixMI \\
    \midrule
    Chloride            & 0.12993 & 0.12549 & 0.10865 & 0.10650 & 0.10575 & 0.10836 & 0.08664 & \bfseries 0.08603 \\
    Potassium           & 0.11533 & 0.11256 & 0.11222 & 0.10963 & 0.10997 & 0.10822 & 0.09453 & \bfseries 0.09442 \\
    Bicarbonate         & 0.13196 & 0.12905 & 0.12371 & 0.12231 & 0.12275 & 0.11984 & 0.10302 & \bfseries 0.10254 \\
    Sodium              & 0.12525 & 0.11939 & 0.11557 & 0.10159 & 0.10138 & 0.10727 & \bfseries 0.08787 & 0.08807 \\
    Hematocrit          & 0.11436 & 0.10780 & 0.09189 & 0.06518 & 0.06558 & 0.06726 & 0.05486 & \bfseries 0.05482 \\
    Hemoglobin          & 0.14168 & 0.12997 & 0.06556 & 0.05774 & 0.05772 & 0.06301 & 0.05117 & \bfseries 0.05103 \\
    MCV                 & 0.14215 & 0.13732 & 0.13798 & 0.13391 & 0.13474 & 0.13340 & \bfseries 0.11634 & 0.11657 \\
    Platelets           & 0.13855 & 0.13087 & 0.14369 & 0.14203 & 0.14236 & 0.12815 & 0.10090 & \bfseries 0.10070 \\
    WBC count           & 0.13963 & 0.13614 & 0.14583 & 0.14026 & 0.14068 & 0.13060 & 0.10934 & \bfseries 0.10913 \\
    RDW                 & 0.14592 & 0.13668 & 0.15938 & 0.15778 & 0.15836 & 0.13897 & 0.11340 & \bfseries 0.11340 \\
    Blood urea nitrogen (BUN) & 0.12358 & 0.12720 & 0.16345 & 0.15160 & 0.15189 & 0.11814 & 0.09479 & \bfseries 0.09410 \\
    Creatinine          & 0.13341 & 0.12803 & 0.14904 & 0.13211 & 0.13212 & 0.12217 & 0.10067 & \bfseries 0.10014 \\
    Glucose             & 0.12491 & 0.12420 & 0.11677 & 0.11771 & 0.11794 & 0.11921 & \bfseries 0.10493 & 0.10501 \\
  \bottomrule
\end{tabular}
\label{tab:maseVar}
\end{table*}

\subsection{The Imputation Model}

The proposed mixture model provides flexibility in changing base models and can be solved under various assumptions. For example, a full mixture model consists of all three base models introduced earlier. However, under the assumption of only linear correlations within time series, mixture models composed of only the two linear components (denoted as MixMI-LL) can be used. Although the choice of mixtures can be determined empirically or based on expert knowledge of the data, we propose an imputation model (denoted as MixMI) that chooses the type of mixture automatically.

In MixMI, for each variable and time index, we train both the MixMI-LL and the full mixture model, then select the one with lower training error as the final mixture model that is used to perform imputation in inference. We use the absolute training error, instead of likelihood, as the selection criterion because the likelihood of a mixture model with two linear models is not of the same scale as the full mixture model.

\section{Experimental Setup} \label{ES}

\subsection{Real-world Datasets}

We collect two real-world datasets from the Medical Information Mart for Intensive Care (MIMIC-III) database \cite{johnson2016mimic} and the Northwestern Medicine Enterprise Data Warehouse (NMEDW). Each dataset contains inpatient test results from 13 laboratory tests. These tests are quantitative, frequently measured on hospital inpatients and commonly used in downstream classification tasks \cite{le1993new,che2018recurrent}. They are the same as those used in \cite{luo20173d} in their imputation study. We organize the data by unique admissions. We distinguish multiple admissions of the same patient. Each admission consists of time series of the 13 laboratory tests. Although our model is evaluated only on continuous variables, they are also applicable on categorical variables with a simple extension (e.g. through predictive mean matching \cite{little1988missing}).  

In both MIMIC-III and NMEDW datasets, the length of time series varies across admissions. To apply our imputation model on these datasets, we truncate time series so that they have the same length. The length is the average number of time points across all admissions. Before truncating, the average number of time points in MIMIC-III dataset is 11. We first exclude admissions that have less than 11 time points, and then we truncate time series by removing measurements taken after the 11-th time point. We also exclude admissions that contain time series that have no observed values. Our MIMIC-III dataset includes 26,154 unique admissions and the missing rate is about 28.71\%. The same data collection procedure is applied on the NMEDW dataset (approved by Northwestern IRB) where we end up with 13,892 unique admissions that have 7 time points. The missing rate of the NMEDW dataset is 24.22\%.

\subsection{Synthetic Datasets}

We create synthetic datasets to explore and discover the properties of the data that might benefit our model and/or comparison models. In synthetic datasets, we augment the correlation between measurements and times. We do not augment correlations by imposing strong constraints on time series where closer measurements have closer values. Instead, we generate synthetic times by altering real times so that the constraints in synthetic data are ``slightly'' stronger than real-world data. We also introduce a scaling factor $d$ to control the strength of the constraints in synthetic data.

The synthetic datasets are generated based on the real-world MIMIC-III dataset. We move two consecutive times of a time series closer, if the relative difference $\Delta \tilde{x}$ in two consecutive measurements is smaller than the relative difference $\Delta \tilde{t}$ in two consecutive times. The relative differences $\Delta \tilde{x}$ and $\Delta \tilde{t}$ of a time series are given by
\begin{displaymath}
\begin{split}
\Delta \tilde{x}_{i} &= \frac{|x_{i} - x_{i-1}|}{\sum_{i=2}^{B} |x_{i} - x_{i-1}|} \\
\Delta \tilde{t}_{i} &= \frac{|t_{i} - t_{i-1}|}{\sum_{i=2}^{B} |t_{i} - t_{i-1}|}.
\end{split}
\end{displaymath}

The scaling factor $d \in (0,1)$ controls how much we move times. If $d=0$, we do not move times. In other words, the synthetic dataset at $d=0$ is the same as the real-world MIMIC-III dataset. As $d$ increases, stronger constraints are introduced to synthetic data. The synthetic time $t'$ for a time series is generated as follows:
\begin{displaymath}
\begin{split}
t'_{i} &= 
\begin{cases}
    t_{1}, & \text{if } i = 1\\
    t_{i} + \sum_{j=2}^{i} [d (\Delta \tilde{x}_{j} - \Delta \tilde{t}_{j}) S], & \text{otherwise}
\end{cases} \\
S &= \sum_{j=2}^{B} (|t_{j} - t_{j-1}|).
\end{split}
\end{displaymath}

If a time series has missing values, we first calculate the synthetic times for the observed measurements. Then we perform a linear interpolation between real times and synthetic times for observed measurements to generate synthetic times for missing measurements.

\subsection{Evaluation of Imputation Quality}

We randomly mask 20\% observed measurements in a data set as missing and treat the masked values as the test set. The remaining observed values are used in training. We impute originally missing and masked values together, and compare the imputed values with the ground truth for masked data to evaluate imputation performance. The Mean Absolute Scaled Error (MASE) \cite{hyndman2006another} is used to measure the quality of imputation on the test set. MASE is a scale-free measure of the accuracy of predictions and recommended by Hyndman et al \cite{hyndman2006another,franses2016note} to measure the accuracy of predictions for series. We calculate MASE for all variables and take a weighted average according to the number of masked values of a variable to get an overall MASE per dataset. 

Let $mask_{p,v}$ be the set of cardinality $I_{p,v}$ of all time indices that have been masked for patient $p$ and variable $v$. Also let $Y_{p,v} = (x^{obs}_{p,v,j})_{j}$ be the sequence of length $J_{p,v}$ of all observed values for patient $p$ and variable $v$, and let $\tilde{x}_{p,v,i}$ represent the imputed value. The MASE for variable $v$ is defined as
\begin{displaymath}
\begin{split}
MASE(v) = \frac{1}{\sum_{\bar{p}} I_{\bar{p},v}} \sum_{p} \frac{\sum_{i \in mask_{p,v}} |\tilde{x}_{p,v,i} - x^{obs}_{p,v,i}|}{\frac{J_{p,v}}{J_{p,v}-1}\sum_{j=2}^{J_{p,v}} |Y_{p,v,j}-Y_{p,v,j-1}|}.
\end{split}
\end{displaymath}

To show the effectiveness of our imputation model MixMI and its variant MixMI-LL, we compare MASE scores of MixMI and MixMI-LL with other six imputation methods: (a) MICE \cite{buuren2011mice} with 100 imputations, where the average of all imputations are used for evaluation; (b) the pure Gaussian processes, where a GP model is fitted to the observed data of each time series using GPfit \cite{macdonald2015gpfit} in R and missing values are replaced with the predictions from the fitted GP models; (c) 3D-MICE, a state-of-the-art imputation model \cite{luo20173d} for which we obtain their code and adapt it to account for our use of the tensor representation; (d) multi-task Gaussian process prediction (MTGP) \cite{bonilla2008multi}; (e) a Gaussian mixture model for imputation (GMM) \cite{di2007imputation,murphy2012machine} and (f) M-RNN, a state-of-the-art RNN-based imputation model \cite{yoon2018estimating}. 

To tune hyperparameters if any in these models, we mask out 20\% observed measurements in the training set as a validation set and tune hyperparameters on the validation set. We introduce $\pi^{(1)}, \pi^{(2)}$ and $\pi^{(3)}$ as hyperparameters of our imputation model. In our experiments, all mixture models start with the same mixing weights, i.e., $\pi^{(k)}_{v,b} = \pi^{(k)}$ for $k=1,2,3$. Other trainable parameters are initialized directly from the data after initial imputation and do not require a manual specification. 


We also evaluate our imputation model under a classification task on our MIMIC dataset and compare it with GRU-D \cite{che2018recurrent}, a state-of-the-art RNN model for classifications with multivariable time series, which does not require a separate data imputation stage. We build a RNN model with the standard Gated Recurrent Unit as our classification model (GRU-MixMI) trained on data imputed by MixMI. GRU-D is trained on data without imputation. We predict whether a patient dies within 30 days after admission, instead of 48 hours as in \cite{che2018recurrent}, because of severe imbalance in our cohort, where only 0.57\% patients have positive labels within 48 hours.

\section{Results} \label{RES}

\subsection{Performance Comparison}

\subsubsection{Imputation task}
\begin{figure}[htbp]
  \centering
  \includegraphics[width=\linewidth]{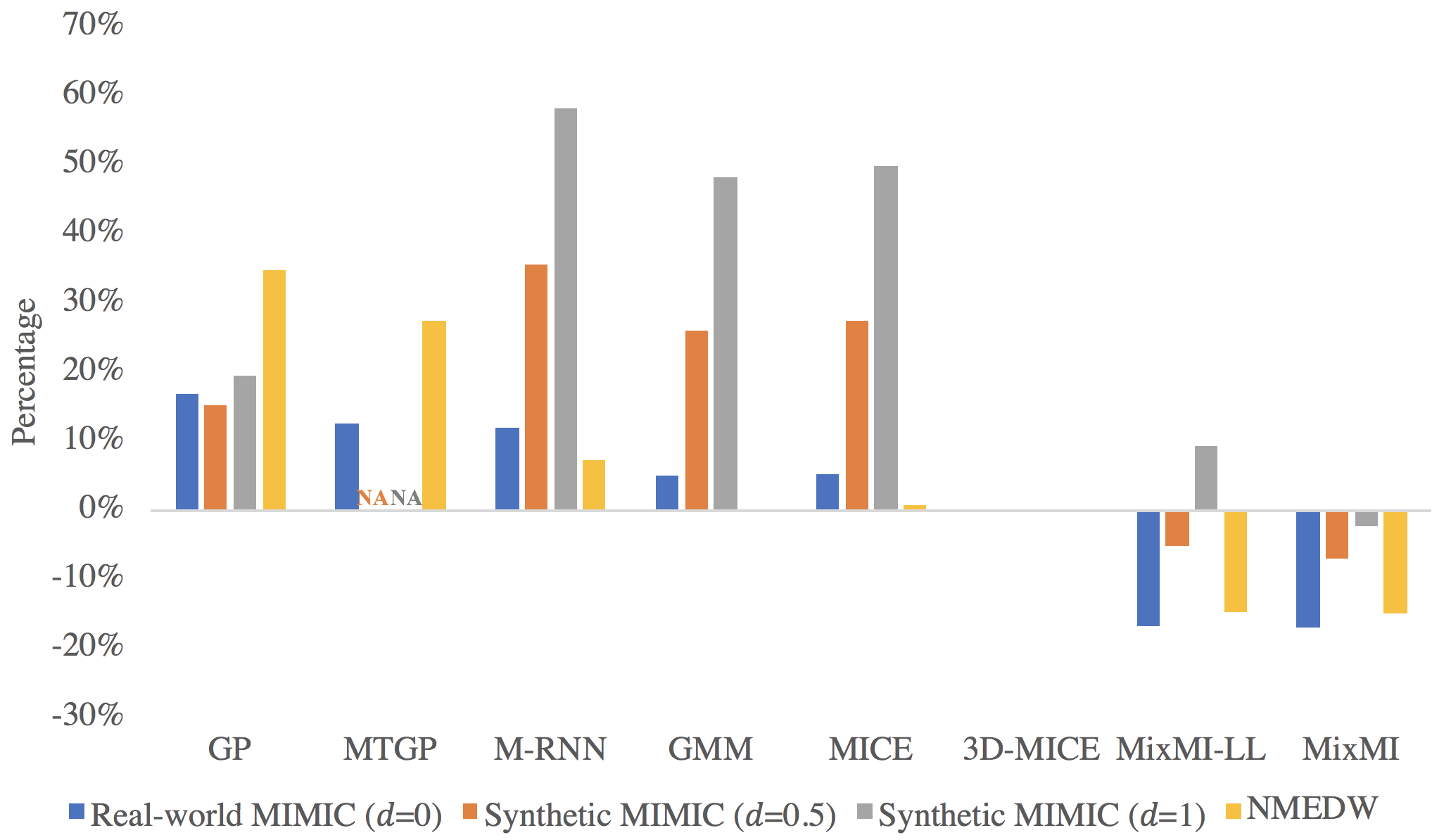}
  \caption{Percentage deviation of MASE score against 3D-MICE}
  \label{fig:mase}
\end{figure}

Table~\ref{tab:maseALL} and Fig.~\ref{fig:mase} compare all imputation models on 4 datasets. Table~\ref{tab:maseALL} shows the overall MASE score of each imputation model on all datasets. Fig.~\ref{fig:mase} provides a comparison for all imputation models in MASE score over 3D-MICE by showing the percentage deviation against 3D-MICE. We select 3D-MICE since it is a state-of-the-art benchmark model. We observe that MixMI outperforms all comparison models on all 4 datasets and is significantly better than the second best model ($p$=.001, permutation test with 1000 replicates) on all 4 datasets.

Table~\ref{tab:maseVar} shows a variable-wise comparison of the imputation models on the real-world MIMIC ($d$=0) dataset. Our two imputation models outperform all comparison models on all variables. MixMI is better than MixMI-LL on most variables. All models except GP and MTGP achieve a much lower error on Hematocrit and Hemoglobin than on other variables. The reason is that these two variables are highly correlated. Those methods that capture the correlation between variables can reasonably infer missing values for Hematocrit from observed measurements of Hemoglobin, and vice versa. Compared to MICE, MixMI achieves even lower errors on these two variables, which indicates that temporal correlations captured by our model help to make better estimation of missing values, even when there is a more dominant cross-sectional correlation.

As shown in Table~\ref{tab:maseALL}, all models except MTGP benefit from the increment of $d$, the scaling factor used to generate synthetic data. The reason is that all models take into account temporal aspects and the measurements in the synthetic time series have stronger temporal correlations as $d$ increases. MICE and GMM also benefit from the temporal correlations because we include time as a feature in addition to variable features. We also try to exclude times, however, experiments show that these two models perform better when times are included. MixMI-LL is outperformed by 3D-MICE when $d$ increases to $1$, while MixMI shows its robustness to the variation of $d$ in our current experimental settings. 

We compare the running times of all models on the real-world MIMIC dataset. MixMI-LL (taking 4.2 hours), GP (1.1 hours), MTGP (1.2 hours) GMM (2.2 hours) and M-RNN (0.9 hours) are the fastest models; 3D-MICE (156.1 hours) is the slowest; MICE (77.5 hours) and MixMI (109.5 hours) are in the middle.

\subsubsection{Classification task}
Our classification model GRU-MixMI is compared with GRU-D on data with different folds and different seeds. First, we make a copy of the original data and split each copy into the same training and test set. One copy is then imputed by MixMI and used to train GRU-MixMI. The other is used by GRU-D directly without imputation. Both models are trained and tested 5 times with different random seeds pertaining to algorithms. The area under the ROC curve (AUC) scores on test sets are reported. This procedure repeats 3 times with different splits. On average, GRU-MixMI ties with GRU-D in terms of AUC scores ($0.7857$ vs. $0.7853$, $p>0.5$). However, compared with models that perform imputation and prediction together, our imputation model, as a pure imputation method, offers more opportunities for other supervised and unsupervised tasks, and modeling choices and effective feature engineering.


\begin{figure}[htbp]
  \centering
  \includegraphics[width=0.9\linewidth]{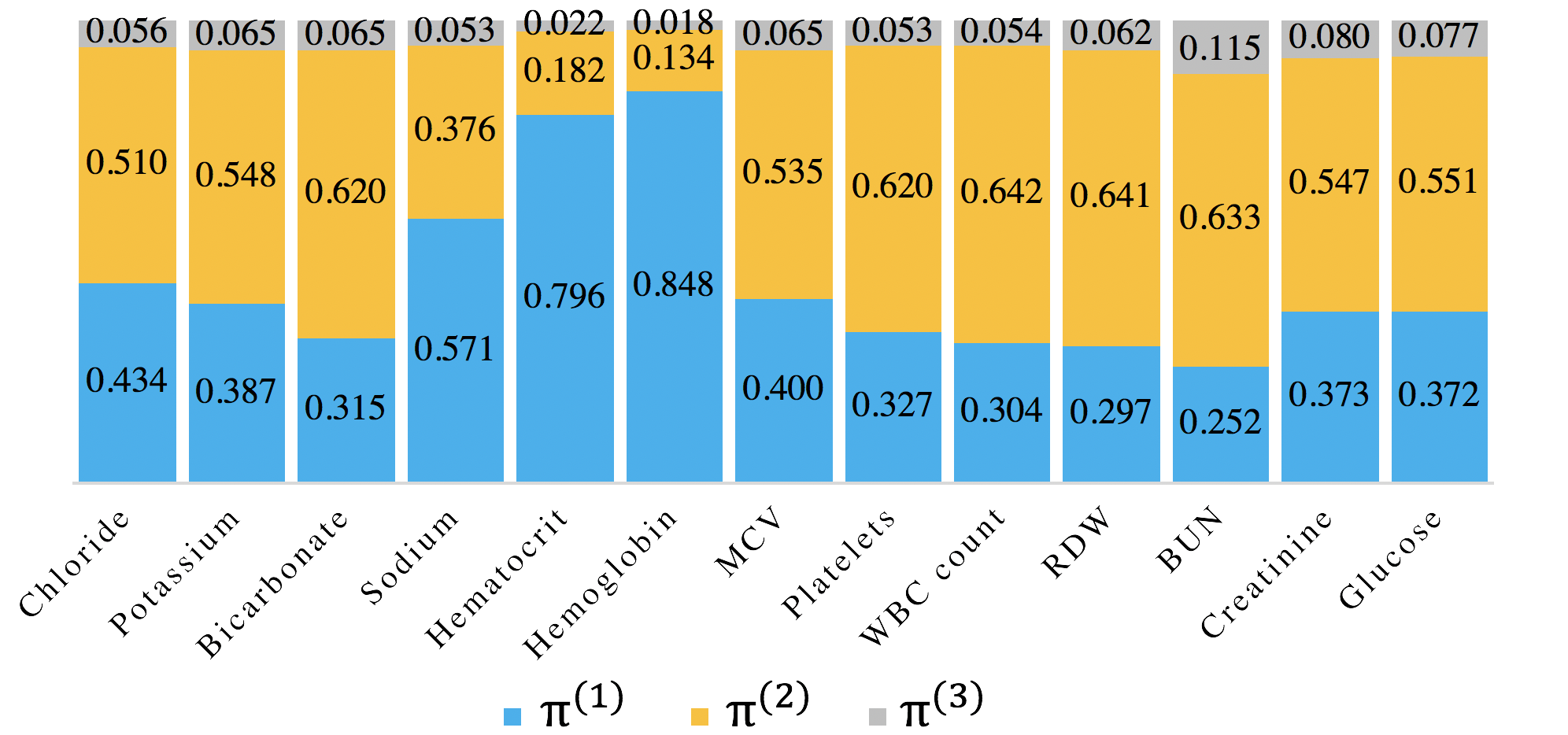}
  \caption{Component weights comparison on real-world MIMIC dataset}
  \label{fig:pis}
\end{figure}

\begin{figure}
    \centering
    \begin{subfigure}{0.45\columnwidth}
        \includegraphics[width=0.9\textwidth]{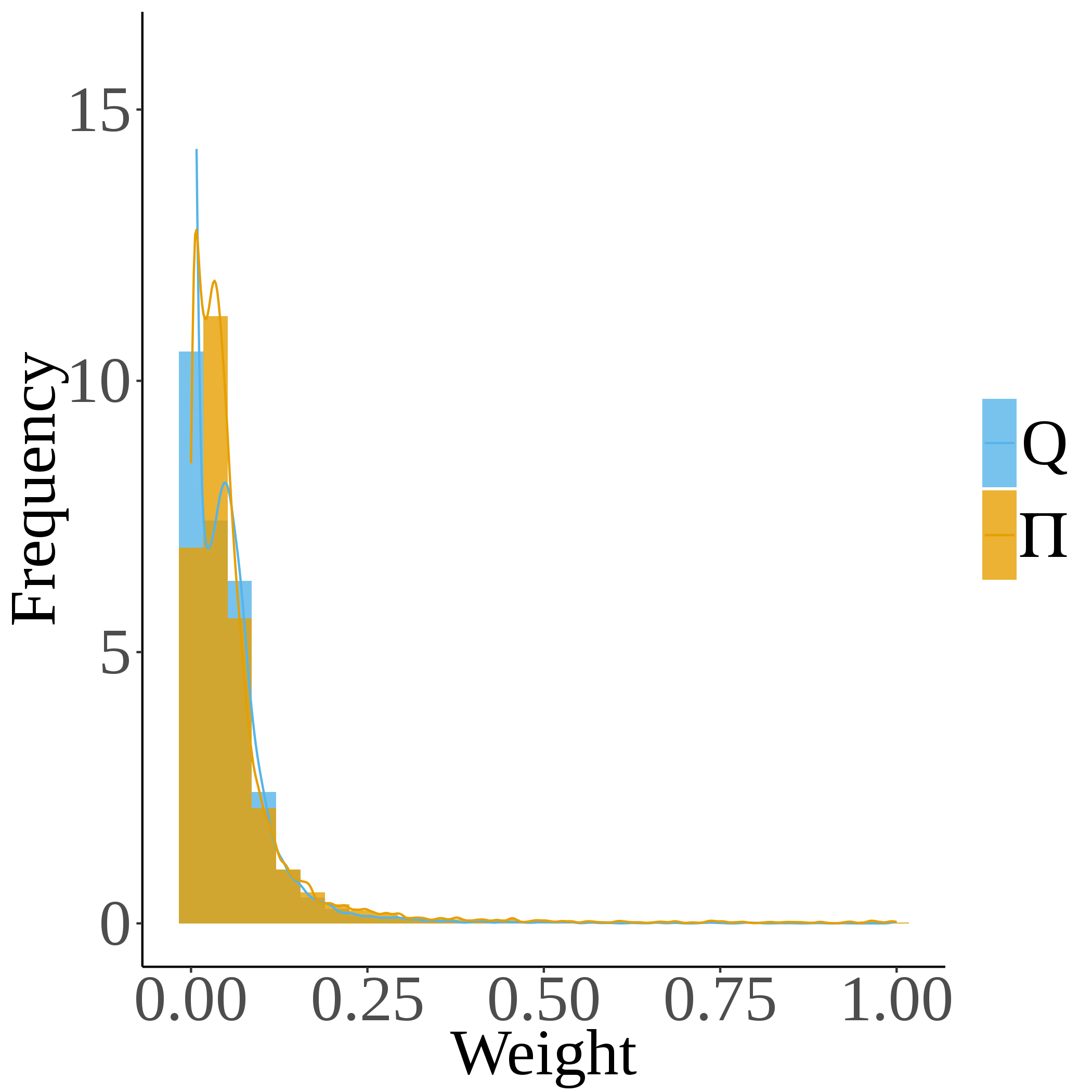}
        \caption{Chloride, T1, $d$=0}
        \label{fig:v2t1a0}
    \end{subfigure}
    \begin{subfigure}{0.45\columnwidth}
        \includegraphics[width=0.9\textwidth]{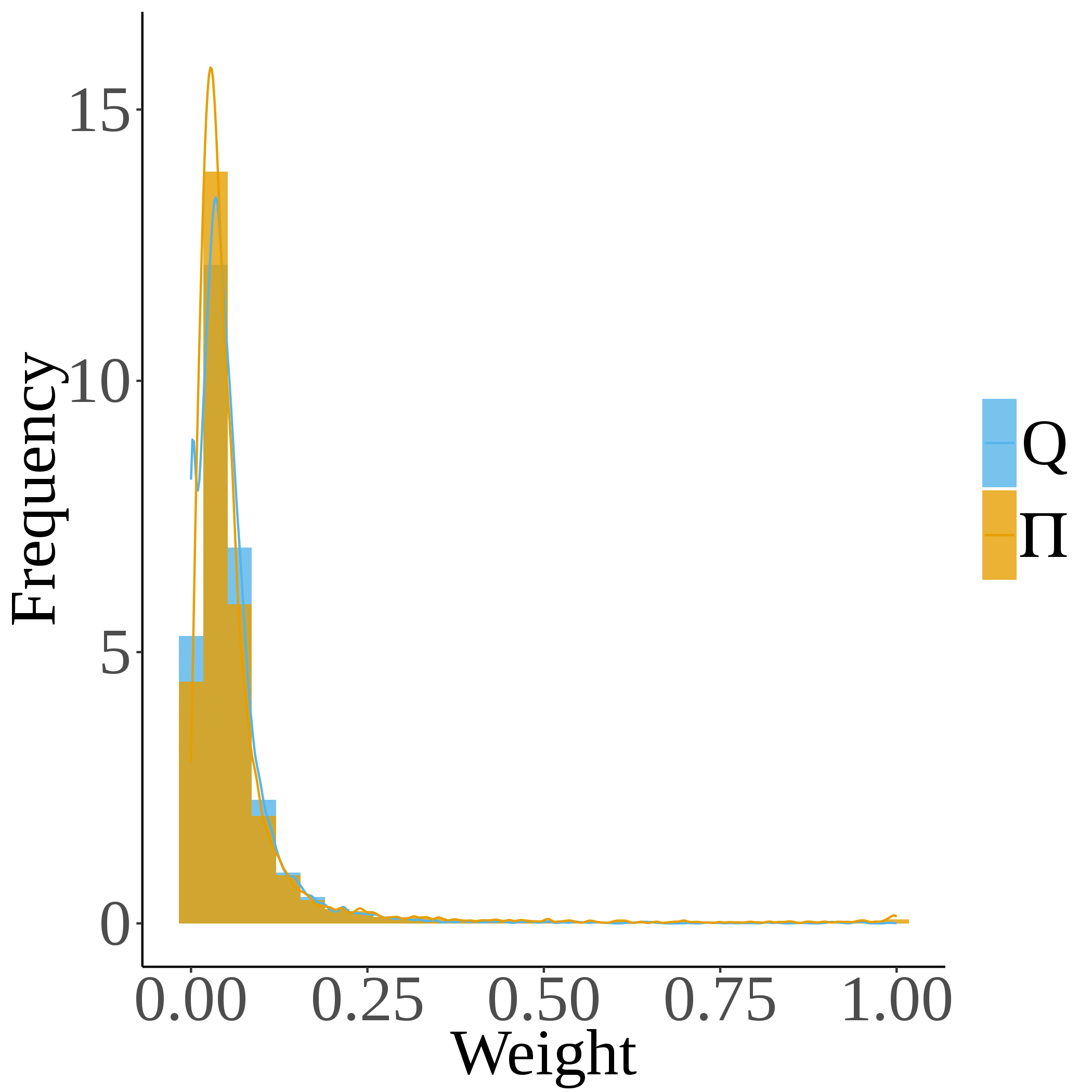}
        \caption{Chloride, T6, $d$=0}
        \label{fig:v2t6a0}
    \end{subfigure}
    \begin{subfigure}{0.45\columnwidth}
        \includegraphics[width=0.9\textwidth]{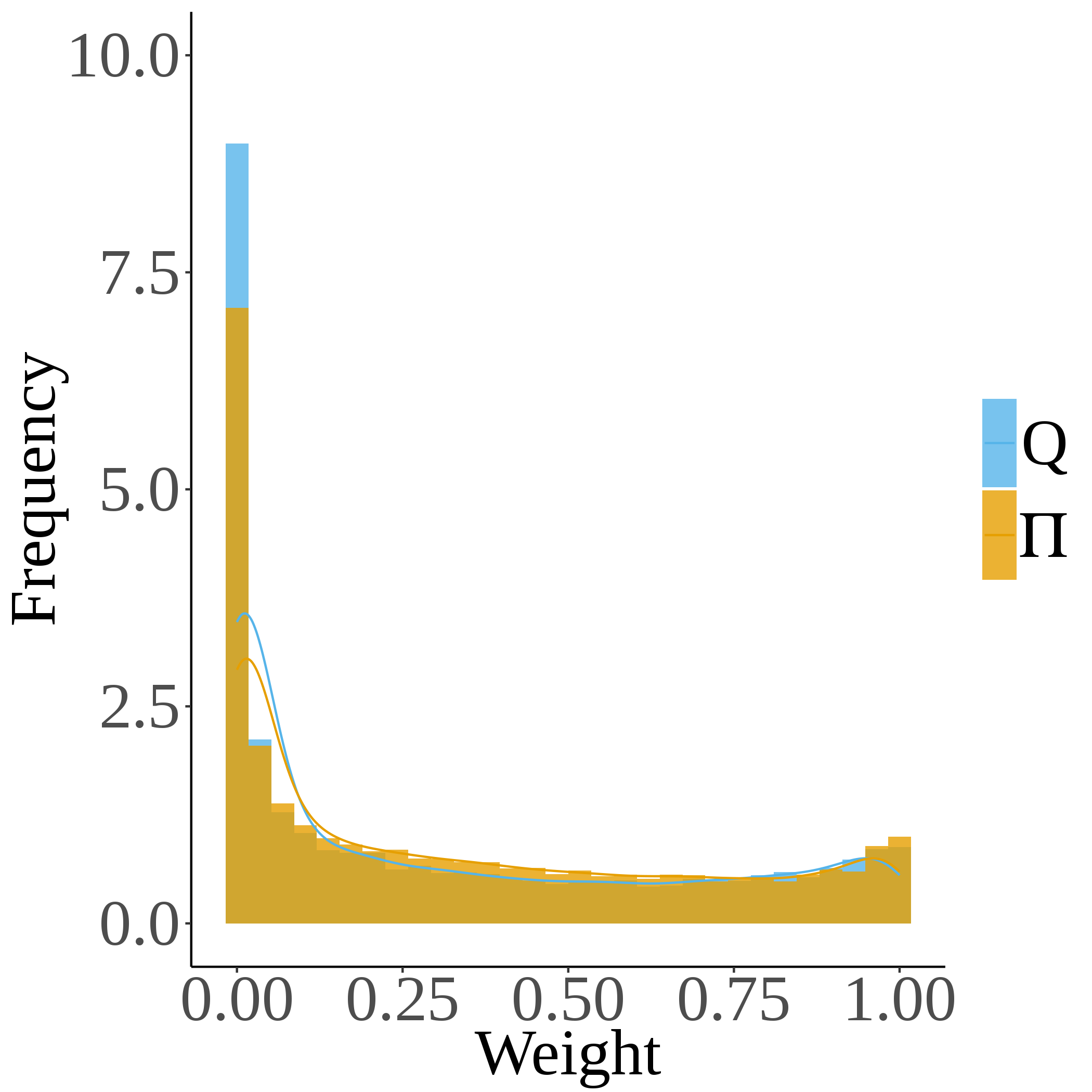}
        \caption{Chloride, T1, $d$=0.5}
        \label{fig:v2t1a0.5}
    \end{subfigure}
    \begin{subfigure}{0.45\columnwidth}
        \includegraphics[width=0.9\textwidth]{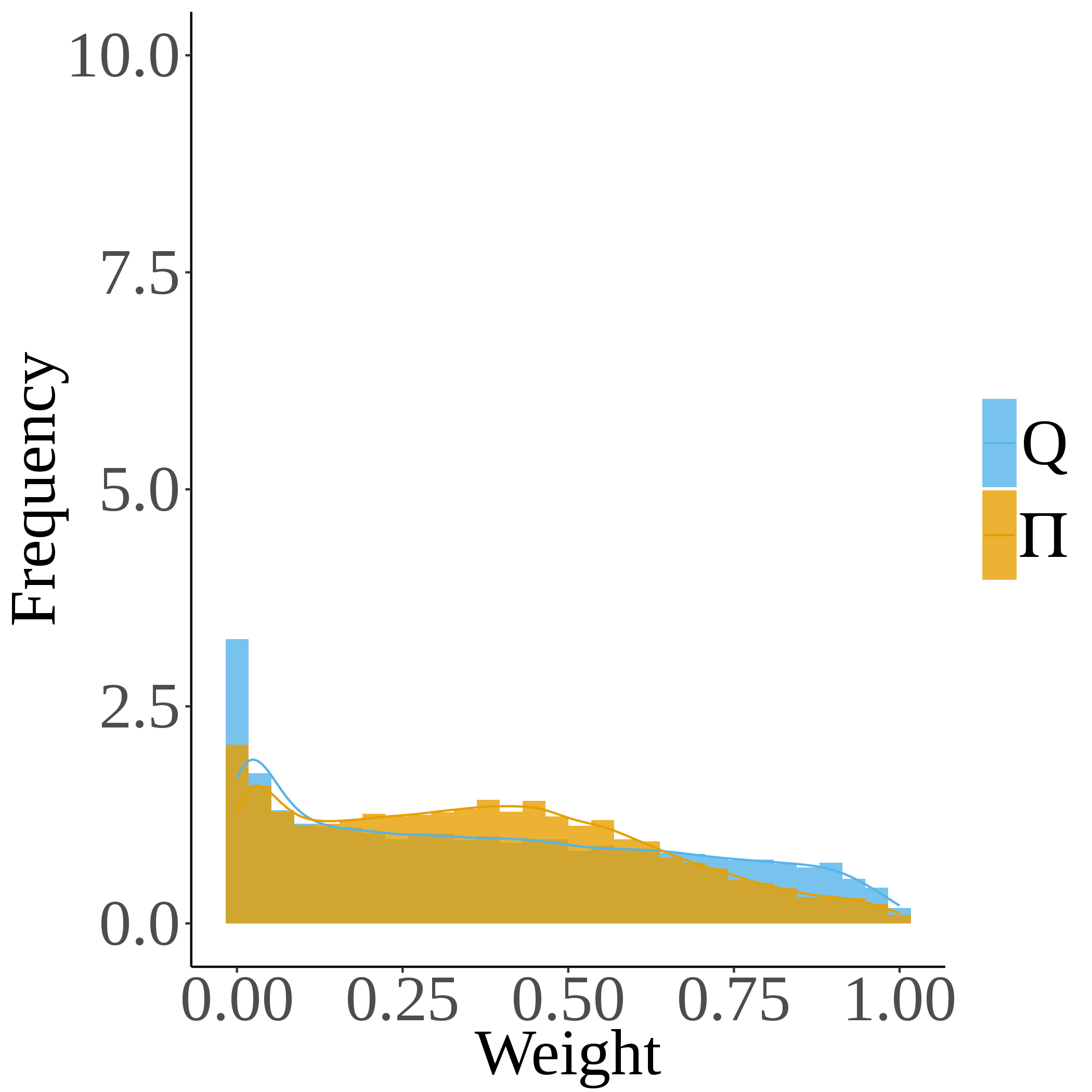}
        \caption{Chloride, T6, $d$=0.5}
        \label{fig:v2t6a0.5}
    \end{subfigure}
    
    \caption{A comparison between individualized mixing weights $\Pi$ and optimized responsibilities $Q$ that GP component should take to ``explain'' observed measurements for training patients. The plots are from the real-world MIMIC ($d$=0) and synthetic MIMIC ($d$=0.5) dataset, and for the mixture models of Chloride at time point 1 and 6. The distributions of the optimized responsibilities are shown in blue and the distributions of individualized mixing weights are in yellow. }\label{fig:weights}
\end{figure}

\subsection{Component Weights}
The key information our imputation model uses to estimate missing values is the component weight $\pi$, which quantifies the interaction of cross-sectional and temporal correlations. We study MixMI on the real-world MIMIC dataset and visualize how the interaction is captured by our model. Fig.~\ref{fig:pis} shows a comparison of all component weights for variables across all patients and times. When imputing Hematocrit and Hemoglobin, our model relies mostly (79.6\% and 84.8\%) on the linear model through the cross-sectional view because of the strong correlation between these two variables. Interestingly, most values of $\pi^{(3)}$ are lower than 10\%, which indicates that temporal correlations are not strong in these variables in our dataset. However, our model is still able to detect the week temporal correlations and utilizes them to improve imputation. Additionally, we observe that when predicting the missing values at the beginning and the end of a time series, our model reasonably uses a lower value of $\pi^{(3)}$ than in the middle. On average, $\pi^{(3)}$ of the first and the last time indices is 13.9\% lower than those in the middle. This is due to the usage of GPs, which usually produce less confident estimates at the end points of series. Furthermore, we observe an increment of $\pi^{(3)}$ as we impose stronger temporal correlations in synthetic datasets, which further validates the ability of our model in capturing such interaction.

\subsection{Individualized Weights}

By introducing individualized (per patient) mixing weights $\Pi$ defined in \eqref{eq:PI}, we improve the performance in MASE score from 0.08351 to 0.07538, an improvement of 9.73\% compared against the model where each mixture component has a fixed weight for all patient cases. The reason that individualized weights are better than fixed weights in our model might be that they better approximate the responsibilities $Q$ defined in \eqref{eq:Q}.

In training, we can optimize the responsibility a component should take to ``explain'' an observed target value $x_{p,v,b}$ for $p \in P^{tr}_{v,b}$. However, when making inference, the responsibility each component should take to ``explain'' a missing value is unknown, because responsibilities depend on observed target values, according to \eqref{eq:Q}. We have to use $\Pi$, the individualized mixing weights, as an approximation of the responsibilities in inference. As defined in \eqref{eq:PI}, $\Pi$ only depends on the inputs, therefore, we can calculate them when making inferences on the test set. 

In a standard mixture model, we could use $\pi^{(k)}_{v,b}$, which is the average of responsibilities of the $k$th component across all training patients, as a fixed weight that the $k$th component should contribute to impute missing values $x^{mis}_{:,v,b}$ for all test patients. However, patient time series can be very different and the confidence of predictions by the GP component can vary largely across different patient cases. A fixed weight can not reflect such variation in prediction confidence.

We shall view an individualized mixing weight as an approximation of how much responsibility a component should take to impute the missing value for a particular patient case. It is tailored for each patient. To visualize how individualized mixing weights help to produce better estimates, in Fig.~\ref{fig:weights}, we plot the distribution of the individualized weights $\Pi$ of the GP component in the training set and compare it with the distribution of the optimized responsibility values $Q$. The responsibilities that the GP component should take can vary a lot in different patient cases, especially on the synthetic dataset, which implies that it is more reasonable for patients to get individualized mixing weights than a fixed weight. We also observe that the individualized mixing weights reasonably mimic the distribution of the optimized responsibilities on the training set. The improvement of our model on the test set attests that the individualized weights approximate the responsibilities better than fixed weights.

\section{Conclusions} \label{CONCLUDE}

We present and demonstrate a mixture-based imputation model MixMI for multivariable clinical time series. Our model can capture both cross-sectional and temporal correlations in time series. We integrate Gaussian processes with mixture models and introduce individualized mixing weights to further improve imputation accuracy. We show that MixMI can provide more accurate imputation than several state-of-the-art imputation models. Although in this work our model is tested on inpatient clinical data, the proposed idea is general and should work for all multivariable time series where interactions of cross-sectional and temporal correlations exist.


\bibliographystyle{IEEEtran}
\bibliography{IEEEabrv,ref}


\clearpage

\appendices

\section{Parameter Estimation in EM} \label{A1}

In the E (Expectation) step, we calculate the responsibilities $w^{(k)}_{p,v,b} = Q_p(q_{v,b}=k)$ for $p \in P^{tr}_{v,b}$ using the current values of the parameters in iteration $j$:
\begin{displaymath}
\begin{split}
[&C^{(1)}_{p,v,b}]^{(j)} \\ 
&= [\pi^{(1)}_{v,b}]^{(j)} [D^{(1)}_{p,v,b}]^{(j)} \mathcal{N} (x_{p,v,b}|x_{p,-v,b}[\beta^{(1)}_{v,b}]^{(j)},[{\sigma^{(1)}_{v,b}}^2]^{(j)}) \\
[&C^{(2)}_{p,v,b}]^{(j)} \\ 
&= [\pi^{(2)}_{v,b}]^{(j)} [D^{(2)}_{p,v,b}]^{(j)} \mathcal{N} (x_{p,v,b}|x_{p,v,-b}[\beta^{(2)}_{v,b}]^{(j)},[{\sigma^{(2)}_{v,b}}^2]^{(j)}) \\
[&C^{(3)}_{p,v,b}]^{(j)} \\ 
&= [\pi^{(3)}_{v,b}]^{(j)} [D^{(3)}_{p,v,b}]^{(j)} \mathcal{N} (x_{p,v,b}|m^G([\alpha_{p,v,b}]^{(j)}),\Sigma^G([\alpha_{p,v,b}]^{(j)})) \\
[&D^{(k)}_{p,v,b}]^{(j)} = \mathcal{N} (V_{p,v,b}|[\boldsymbol{\mu}^{(k)}_{v,b}]^{(j)},[\boldsymbol{\Sigma}^{(k)}_{v,b}]^{(j)}), k = 1,2,3 \\
[&w^{(k)}_{p,v,b}]^{(j)} = \frac{[C^{(k)}_{p,v,b}]^{(j)}}{\sum_{i=1}^{3} [C^{(i)}_{p,v,b}]^{(j)}}, k = 1,2,3
\end{split}
\end{displaymath}

Let $Z_{v,b} = (x_{p,-v,b})_{p \in P^{tr}_{v,b}}$ and $Y_{v,b} = (x_{p,v,-b})_{p \in P^{tr}_{v,b}}$. In the M (Maximization) step, we re-estimate the parameters in iteration $(j+1)$ using the $j$th responsibilities:
\begin{displaymath}
\begin{split}
[\pi^{(k)}_{v,b}]^{(j+1)} &= \frac{1}{|P^{tr}_{v,b}|} \sum_{p \in P^{tr}_{v,b}} [w^{(k)}_{p,v,b}]^{(j)}, k = 1,2,3 \\
[\boldsymbol{\mu}^{(k)}_{v,b}]^{(j+1)} &= \frac{\sum_{p \in P^{tr}_{v,b}} [w^{(k)}_{p,v,b}]^{(j)} V_{p,v,b}}{\sum_{p \in P^{tr}_{v,b}} [w^{(k)}_{p,v,b}]^{(j)}} \\
[\boldsymbol{\Sigma}^{(k)}_{v,b}]^{(j+1)} &= \frac{1}{\sum_{p \in P^{tr}_{v,b}} [w^{(k)}_{p,v,b}]^{(j)}} \sum_{p \in P^{tr}_{v,b}} [w^{(k)}_{p,v,b}]^{(j)} [U^{(k)}_{p,v,b}]^{(j+1)} \\
[U^{(k)}_{p,v,b}]^{(j+1)} &= \{V_{p,v,b} - [\boldsymbol{\mu}^{(k)}_{v,b}]^{(j+1)}\}\{V_{p,v,b}-[\boldsymbol{\mu}^{(k)}_{v,b}]^{(j+1)}\}' \\ 
[\beta^{(1)}_{v,b}]^{(j+1)} &= \{\{Z'_{v,b}[\mathbf{w}^{(1)}_{v,b}]^{(j)} Z_{v,b}\}^{-1} Z'_{v,b} [\mathbf{w}^{(1)}_{v,b}]^{(j)} x^{obs}_{:,v,b}\}' \\
[{\sigma^{(1)}_{v,b}}^2]^{(j+1)} &= \frac{1}{\sum_{p \in P^{tr}_{v,b}} [w^{(1)}_{p,v,b}]^{(j)}} [S^{(1)}_{v,b}]^{(j+1)} \\
[S^{(1)}_{v,b}]^{(j+1)} &= \sum_{p \in P^{tr}_{v,b}} [w^{(1)}_{p,v,b}]^{(j)} \{x_{p,v,b} - x_{p,-v,b} [\beta^{(1)}_{v,b}]^{(j+1)}\}^2 \\
[\beta^{(2)}_{v,b}]^{(j+1)} &= \{\{Y'_{v,b}[\mathbf{w}^{(2)}_{v,b}]^{(j)} Y_{v,b}\}^{-1} Y'_{v,b} [\mathbf{w}^{(2)}_{v,b}]^{(j)} x^{obs}_{:,v,b}\}' \\
[{\sigma^{(2)}_{v,b}}^2]^{(j+1)} &= \frac{1}{\sum_{p \in P^{tr}_{v,b}} [w^{(2)}_{p,v,b}]^{(j)}} [S^{(2)}_{v,b}]^{(j+1)} \\
[S^{(2)}_{v,b}]^{(j+1)} &= \sum_{p \in P^{tr}_{v,b}} [w^{(2)}_{p,v,b}]^{(j)} \{x_{p,v,b} - x_{p,v,-b} [\beta^{(2)}_{v,b}]^{(j+1)}\}^2 \\
[\theta_{v,b}]^{(j+1)} &= G([\mathbf{w}^{(3)}_{v,b}]^{(j)},[\theta_{v,b}]^{(j)},x^{obs}_{:,v,b},Y_{v,b},t_{:,v,:})
\end{split}
\end{displaymath}
where $[\mathbf{w}^{(k)}_{v,b}]^{(j)}$ is the vector of $[w^{(k)}_{p,v,b}]^{(j)}$ for $p \in P^{tr}_{v,b}$ in iteration $j$. The kernel parameters $\theta_{v,b}$ of GP models are evaluated by function $G$, a gradient descent method that calculates the estimates of $[\theta_{v,b}]^{(j+1)}$ to maximize $\mathcal{L}_{v,b} (\gamma)$, using $[\theta_{v,b}]^{(j)}$ as the starting point. The first order derivatives of $\mathcal{L}_{v,b} (\gamma)$ with respect to $\theta_{v,b}$ that are used in $G$ are given in Appendix C.

\section{GP Model} \label{A2}

We assume the GP model discussed here in a mixture model for a certain variable and time, and thus we exclude the subscripts $v$ and $b$. We use $x_{p,t}$ to denote a measurement of the time series $x_{p}$ at time $t$ for patient $p$ of a certain variable. We use $x_{p,-t}$ to denote a time series without the measurement at time $t$. The GP model is given by
\begin{displaymath}
\begin{split}
x_{p,t} &= \mu_{p,t} + f(t), \\
f(t) &\sim \mathcal{GP} (0,\mathcal{K}(t,t'))
\end{split}
\end{displaymath}
where $\mu_{p,t}$ is the overall mean of the model and $f(t)$ is a Gaussian process with mean of $0$ and covariance of $\mathcal{K}(t,t')$.
Following the maximum likelihood approach, the best linear unbiased predictor (BLUP) \footnote{Sacks, Jerome, et al. "Design and analysis of computer experiments." Statistical science (1989): 409-423.} at $t$ and the mean squared error are
\begin{displaymath}
\begin{split}
m^{(3)}(\theta, x_{p,-t}, \bar{t}) &= (\frac{1-r^T R^{-1} 1_n}{1_n^T R^{-1}1_n}1_n^T + r^T)R^{-1}x_{p,-t} \\
\Sigma^{(3)}(\theta, x_{p,-t}, \bar{t}) &= \sigma_f^2[1-r^T R^{-1}r + \frac{(1-1_n^T R^{-1} r)^2}{1_n R^{-1} 1_n}]
\end{split}
\end{displaymath}
where $r_t(t')=corr(f(t),f(t'))$, $r$ is the vector of $r_t(t')$ for all possible $t$, $\bar{t}$ is a vector of time except for time $t$, $R$ is the $(B-1) \times (B-1)$ correlation matrix and the correlation function is given by
\begin{displaymath}
R_{t,t'} = \exp(-\theta |t - t'|^2).
\end{displaymath}

The estimator $\sigma^2$ is given by
\begin{displaymath}
\sigma_f^2 = \frac{C^T R^{-1} C}{n}, C = x_{p,-t} - 1_n (1_n^T R^{-1} 1_n)^{-1}(1_n^T R^{-1} x_{p,-t})
\end{displaymath}
where $1_n$ is a vector with length $(B-1)$ of all ones.

\newcommand{\Rinv}{R^{-1}}
\newcommand{\mIn}{1_n}
\newcommand{\ITRI}{\mIn^{T} \Rinv \mIn}
\newcommand{\Rdl}{\frac{\partial R}{\partial \theta}}
\newcommand{\Rinvdl}{\frac{\partial R^{-1}}{\partial \theta}}
\newcommand{\ITRIdl}{\mIn^{T} \Rinvdl \mIn}
\newcommand{\rT}{r}
\newcommand{\rTdl}{\frac{\partial r}{\partial \theta}}
\newcommand{\rdl}{\frac{\partial r^{T}}{\partial \theta}}
\newcommand{\rTRinvI}{(\rT \Rinv \mIn)}
\newcommand{\rTRinvIdl}{(\rTdl \Rinv + \rT \Rinvdl) \mIn}
\newcommand{\fc}{[1-\rTRinvI]}
\newcommand{\gc}{\ITRI}
\newcommand{\fcdl}{-\rTRinvIdl}
\newcommand{\gcdl}{\ITRIdl}
\newcommand{\partI}{\frac{\fc}{\gc} \mIn^{T} + \rT}
\newcommand{\partIdl}{\frac{\fcdl (\gc)}{\gc^2} \\ &- \frac{(\gcdl) \fc}{\gc^2} \mIn^{T} + \rTdl}
\newcommand{\yhatdl}{((\partIdl \Rinv + (\partI) \Rinvdl) x_{p,-t})}
\newcommand{\rTRinvr}{(\rT \Rinv r^{T})}
\newcommand{\rTRinvrdl}{(\rTdl \Rinv r^{T} + \rT \Rinvdl r^{T} + \rT \Rinv \rdl)}
\newcommand{\fcI}{(1-\mIn^{T} \Rinv r^{T})^{2}}
\newcommand{\fcIdl}{2 (1-\mIn^{T} \Rinv r^{T}) [- \mIn^{T} (\Rinvdl r^{T} + \Rinv \rdl)]}
\newcommand{\gcI}{\ITRI}
\newcommand{\gcIdl}{\ITRIdl}
\newcommand{\partII}{\frac{\fcI}{\gcI}}
\newcommand{\partIII}{1-\rTRinvr + \partII}
\newcommand{\ITRY}{(\mIn^{T} \Rinv x_{p,-t})}
\newcommand{\ITRYdl}{(\mIn^{T} \Rinvdl x_{p,-t})}
\newcommand{\partIV}{x_{p,-t} - \mIn \frac{\ITRY}{\ITRI}}
\newcommand{\partIVdl}{- \mIn \frac{1}{(\ITRI)^{2}} [\ITRYdl (\ITRI) \\ & - (\ITRIdl) \ITRY]} 
\newcommand{\sigmaII}{[\frac{1}{n} (\partIV)^{T} \Rinv (\partIV)]}
\newcommand{\sigmaIIdl}{[\frac{1}{n} (\partIVdl^{T} \Rinv (\partIV) + (\partIV)^{T} \Rinvdl (\partIV) + (\partIV)^{T} \Rinv \partIVdl)]}
\newcommand{\sII}{[\sigmaII (\partIII)]}
\newcommand{\sIIdl}{[\sigmaII \partIIIdl + \sigmaIIdl (\partIII)]}
\newcommand{\h}{\sII}
\newcommand{\fdl}{\sIIdl}
\newcommand{\g}{((\partI \Rinv)x_{p,-t})}
\newcommand{\gdl}{\yhatdl}
\newcommand{\Ldl}{[w_{p} (-\frac{1}{2\h} \fdl + \frac{(s_{p} - \g) \gdl}{\h} + \frac{\fd (s_{p} - \g)^2}{2 \h^2})]}

\section{Partial Derivatives in GP} \label{A3}

\noindent To simplify the notations, we assume that the likelihood function $L$ under consideration is for a mixture model for a certain variable and time. The partial derivative with respect to Gaussian process parameters $\theta$ is
\begin{displaymath}
\frac{\partial L}{\partial \theta}  = \sum_{p=1}^{|p^{tr}|} w_{p} \frac{\partial}{\partial \theta} \ln \mathcal{N} (x_{p,t};m^{(3)}(\theta,x_{p,-t},\bar{t}),\Sigma^{(3)}(\theta,x_{p,-t},\bar{t})).
\end{displaymath}

Letting $g_{p}(\theta) = m^{(3)}(\theta,x_{p,-t},\bar{t})$ and $h_{p}(\theta) = \Sigma^{(3)}(\theta,x_{p,-t},\bar{t})$, we have
\begin{displaymath}
\begin{split}
\frac{\partial L}{\partial \theta} &= \sum_{p=1}^{|p^{tr}|} w_{p} \frac{\partial}{\partial \theta} \ln \mathcal{N} (x_{p,t};g_{p}(\theta),h_{p}(\theta)) \\
&= \sum_{p=1}^{|p^{tr}|} w_{p} \frac{\partial}{\partial \theta} \{\ln \frac{1}{\sqrt[]{2 \pi h_{p}(\theta)}} - \frac{[x_{p,t}-g_{p}(\theta)]^2}{2 h_{p}(\theta)}\} \\
&= \sum_{p=1}^{|p^{tr}|} w_{p} \{-\frac{1}{2 h_{p}(\theta)} \frac{\partial h_{p}(\theta)}{\partial \theta} - \frac{\partial}{\partial \theta} \frac{[x_{p,t} - g_{p}(\theta)]^2}{2 h_{p}(\theta)}\} \\
&= \sum_{p=1}^{|p^{tr}|} w_{p} \{-\frac{1}{2 h_{p}(\theta)} \frac{\partial h_{p}(\theta)}{\partial \theta} \\ 
&- \frac{1}{2 h^2_{p}(\theta)} \{2[x_{p,t}-g_{p}(\theta)][-\frac{\partial g_{p}(\theta)}{\partial \theta}]h_{p}(\theta) \\ & - \frac{\partial h_{p}(\theta)}{\partial \theta} [x_{p,t}-g_{p}(\theta)]^2\}\} \\
&= \sum_{p=1}^{|p^{tr}|} w_{p} \{-\frac{1}{2 h_{p}(\theta)} \frac{\partial h_{p}(\theta)}{\partial \theta} \\ 
&+ \frac{[x_{p,t}-g_{p}(\theta)]\frac{\partial g_{p}(\theta)}{\partial \theta}}{h_{p}(\theta)} + \frac{\frac{\partial h_{p}(\theta)}{\partial \theta} [x_{p,t}-g_{p}(\theta)]^2}{2 h^2_{p}(\theta)}\} .
\end{split}
\end{displaymath}

Then $\frac{\partial g_{p}(\theta)}{\partial \theta}$ and $\frac{\partial h_{p}(\theta)}{\partial \theta}$ are given by
\begin{displaymath}
\begin{split}
\frac{\partial g_{p}(\theta)}{\partial \theta} &= (\frac{\partial H_{1}}{\partial \theta} \Rinv + H_{1} \Rinvdl) x_{p,-t} \\
\frac{\partial h_{p}(\theta)}{\partial \theta} &= \sigma_f^2 \frac{\partial H_{3}}{\partial \theta} + \frac{\partial \sigma_f^2}{\partial \theta} H_{3}
\end{split}
\end{displaymath}
where $H_{1}$, $\frac{\partial H_{1}}{\partial \theta}$, $H_{3}$ and $\frac{\partial H_{3}}{\partial \theta}$ are given as follows:
\begin{displaymath}
\begin{split}
H_{1} &= \partI \\
\frac{\partial H_{1}}{\partial \theta} &= \partIdl \\
fc &= \fcI \\
gc &= \gcI \\
\frac{\partial fc}{\partial \theta} &= \fcIdl \\
\frac{\partial gc}{\partial \theta} &= \gcIdl \\
H_{2} &= \partII \\
\frac{\partial H_{2}}{\partial \theta} &= \frac{\frac{\partial fc}{\partial \theta} gc - \frac{\partial gc}{\partial \theta} fc}{gc^2} 
\end{split}
\end{displaymath}

\begin{displaymath}
\begin{split}
H_{3} &= 1-\rTRinvr + H_{2} \\
\frac{\partial H_{3}}{\partial \theta} &= - \rTRinvrdl + \frac{\partial H_{2}}{\partial \theta} \\
H_{4} &= \partIV \\
\frac{\partial H_{4}}{\partial \theta} &= \partIVdl \\
\frac{\partial \sigma^2_f}{\partial \theta} &= \frac{1}{n} [(\frac{\partial H_{4}}{\partial \theta})^{T} \Rinv H_{4} + H_{4}^{T} \frac{\partial R^{-1}}{\partial \theta} H_{4} + H_{4}^{T} \Rinv \frac{\partial H_{4}}{\partial \theta}].
\end{split}
\end{displaymath}


\end{document}